\documentclass[10pt,twocolumn,letterpaper]{article}

\usepackage{cvpr}
\usepackage{times}
\usepackage{epsfig}
\usepackage{graphicx}
\usepackage{times}  
\usepackage{helvet} 
\usepackage{courier}  
\usepackage[hyphens]{url}  
\usepackage{graphicx} 
\usepackage{graphicx}  
\usepackage{caption} 
\usepackage{multirow}
\usepackage{amsthm,amsmath,amssymb}
\usepackage{mathrsfs}
\usepackage{bm}
\usepackage[switch]{lineno}

\usepackage[breaklinks=true,bookmarks=false]{hyperref}

\cvprfinalcopy 

\def\cvprPaperID{****} 

\begin{document}

\title{SDTP: Semantic-aware Decoupled Transformer Pyramid for Dense Image Prediction}

\author{Zekun Li\qquad Yufan Liu\qquad Bing Li\qquad Weiming Hu\qquad Kebin Wu\qquad Pei Wang\\
{\tt\small (lizekun2018,yufan.liu)@ia.ac.cn,
(bli,wmhu)@nlpr.ia.ac.cn,
wukebin@huawei.com,
wangpei@cert.org.cn}
}
\maketitle

\begin{abstract}
   Although transformer has achieved great progress on computer vision tasks, the scale variation in dense image prediction is still the key challenge. Few effective multi-scale techniques are applied in transformer and there are two main limitations in the current methods. On one hand, self-attention module in vanilla transformer fails to sufficiently exploit the diversity of semantic information because of its rigid mechanism. On the other hand, it is hard to build attention and interaction among different levels due to the heavy computational burden. To alleviate this problem, we first revisit multi-scale problem in dense prediction, verifying the significance of diverse semantic representation and multi-scale interaction, and exploring the adaptation of transformer to pyramidal structure. Inspired by these findings, we propose a novel Semantic-aware Decoupled Transformer Pyramid (SDTP) for dense image prediction, consisting of Intra-level Semantic Promotion (ISP), Cross-level Decoupled Interaction (CDI) and Attention Refinement Function (ARF). ISP explores the semantic diversity in different receptive space. CDI builds the global attention and interaction among different levels in decoupled space which also solves the problem of heavy computation. Besides, ARF is further added to refine the attention in transformer. Experimental results demonstrate the validity and generality of the proposed method, which outperforms the state-of-the-art by a significant margin in dense image prediction tasks. Furthermore, the proposed components are all plug-and-play, which can be embedded in other methods.   
\end{abstract}

\section{Introduction}
Inspired by the great success of transformer \cite{vaswani2017attention} in natural language processing (NLP) \cite{devlin2018bert}, there is an increasing effort on applying it to computer vision. Following vision transformer (ViT) \cite{dosovitskiy2020image}, which is the first attempt of applyting transformer to vision, plenty of studies \cite{wang2021pyramid, liu2021swin} have adopted transformer to dense image prediction tasks, such as object detection, semantic segmentation and instance segmentation.

However, the scale variation in dense image prediction is still the key challenge, even for transformer-based methods. While the pyramidal structure represented by Feature Pyramid Network (FPN) \cite{lin2017feature} is an effective method to tackle this problem, but few attempts are made to apply multi-scale technique in transformer. On one hand, vanilla transformer fails to explore the diversity of high-level feature's semantic information, because of the fixed receptive field of patches and the limitations of the self-attention mechanism. On the other hand, establishing attention directly on low-level future maps, which has large size in general, is infeasible due to quadratic computational cost, let alone building attention and interaction among different levels. 

To alleviate this problem, we first revisit the multi-scale problem in dense prediction and obtain several findings. Through the decomposition experiments of FPN, we find that the semantic level (i.e., $C_5$) plays a significant role in dense prediction, and even the single input of $C_5$ can achieve comparable performance. Besides the semantic information, multi-level interaction is indispensable, which can promote the mutual learning among different levels and suppress the redundant information in low levels. In addition, we make various attempts to adapt transformer to FPN. The results show that in the proposed decoupled space, transformer saves large computation and achieves high performance.

Based on these findings, we propose a novel Semantic-aware Decoupled Transformer Pyramid (SDTP) for dense image prediction consisting of three components: Intra-level Semantic Promotion (ISP), Cross-level Decoupled Interaction (CDI) and Attention Refinement Function (ARF). ISP exploits the semantic diversity in various receptive space to amply mine the semantic information, integrating local to global information in transformer flexibly. CDI builds the global attention and interaction across different feature levels with the help of proposed decoupled technique, which also solves the problem of heavy computation. Besides, ARF module is further embedded in attention module in transformer to refine the attention map. These three components are all plugged and can be embedded in various methods.

In summary, this work makes the following contributions:
\begin{itemize}
	\item The multi-scale problem in dense image prediction is revisited with insightful findings of key factors leading to the success of multi-scale prediction: sufficient semantic information and effective interaction computation. Besides, one simple yet effective solution is presented to reduce the computational cost of multi-scale transformer.
	\item We propose SDTP to alleviate scale variation problem in dense image prediction, which is an enhanced drop-in replacement of FPN with transformer, for generating more representative multi-scale features. 
	\item SDTP consists of three components: Intra-level Semantic Promotion (ISP), Cross-level Decoupled Interaction (CDI) and Attention Refinement Function (ARF). ISP makes full use of semantic information in various receptive space. CDI builds a global attention and interaction among different levels in decoupled space, and ARF is embedded in transformer block to refine the attention map. Each of these components can be separately utilized in various methods.
\end{itemize}

\section{Related Works}
\subsection{Dense image prediction task}
\noindent{\textbf{Object detection.}} Most detectors can be divided into two types: one-stage detectors (e.g., RetinaNet \cite{lin2017focal}, OneNet \cite{sun2020onenet}) and multi-stage detectors (e.g., Faster R-CNN \cite{ren2015faster}, Cascade R-CNN \cite{cai2018cascade}). Recently, inspired by the excellent performance of transformer in NLP tasks, some studies applied transformer to object detection tasks. DETR \cite{carion2020end} is proposed to utilize transformer for end-to-end detection for the first time. Deformable DETR \cite{zhu2020deformable} further embeds an effective attention module, leading to better performance than DETR. 

\noindent{\textbf{Segmentation segmentation.}} FCN \cite{long2015fully} utilizes a fully convolutional network to obtain segmentation maps. Inspired by FCN, U-Net \cite{ronneberger2015u} is widely used in medical segmentation, fusing multi-level information for prediction. Besides, the series of DeepLab \cite{liu2019auto, chen2017deeplab} apply dilated convolution to obtain large receptive fields for more spatial information.   

\noindent{\textbf{Instance segmentation.}} By adding a paralleled mask head, Mask R-CNN \cite{he2017mask} extends Faster R-CNN for instance segmentation task. To refine the edge of instances, PointRend \cite{kirillov2020pointrend} provides a more sophisticated mask heads with a set of points. In HTC \cite{chen2019hybrid}, apart from a semantic segmentation branch for contextual information, a cascade structure is used to combine the object detection and instance segmentation for multi-stage prediction. All these dense image prediction tasks are faced with scale variation problem and need multi-scale features for precise prediction.

\subsection{Method for scale variation}
Scale variation of object instances is a giant obstacle in dense image prediction tasks and most current studies rely on multi-scale technique. Feature pyramid network (FPN) \cite{lin2017feature} is a classical structure that contains a top-down pathway to fuse the adjacent features. After that, a series of studies are proposed to further improve FPN. For example, PANet \cite{liu2018path} introduces a bottom-up pathway to shorten information path among different levels and FPG \cite{chen2020feature} utilizes a deep multi-pathway feature pyramid to make the feature fusion in various directions. Different from FPN and its variants, YOLOF \cite{chen2021you} is proposed to utilize a single level feature to alleviate the scale variation through multi-receptive field blocks. Therefore, how to deal with multi-scale and multi-receptive field features effectively is the key for scale variation.

\subsection{Vision transformer}
Recently, a resurgence of work in transformer has led to major advances in vision tasks. ViT \cite{dosovitskiy2020image} constructs a pure transformer backbone for image classification. Following ViT, a series of works were presented. For example, T2T-ViT \cite{yang2019reppoints} splits the image into tokens of overlapping patches, to strengthen the interaction of tokens. In addition, the shifted windows for self-attention are proposed in Swin transformer \cite{liu2021swin}, providing local connections among different windows. Besides the methods above, some other methods explore to integrating CNN and transformer, taking advantage of both sides. For instance, PVT \cite{wang2021pyramid} applies the transformer to the pyramid structure used in ResNet and CVT \cite{wu2021cvt} introduces the convolutions to vision transformer which combines the preponderance of each others. However, most of previous studies concentrate on building the attention in the same scales of backbone, ignoring multi-scale interaction among different levels. Contrastively, the proposed method designs a multi-scale transformer, which can build long-range relationship among different levels to alleviate scale variation in dense prediction tasks.

\section{Revisiting dense multi-scale prediction}
\begin{figure}[!t]
	\centering
	\includegraphics[scale=1]{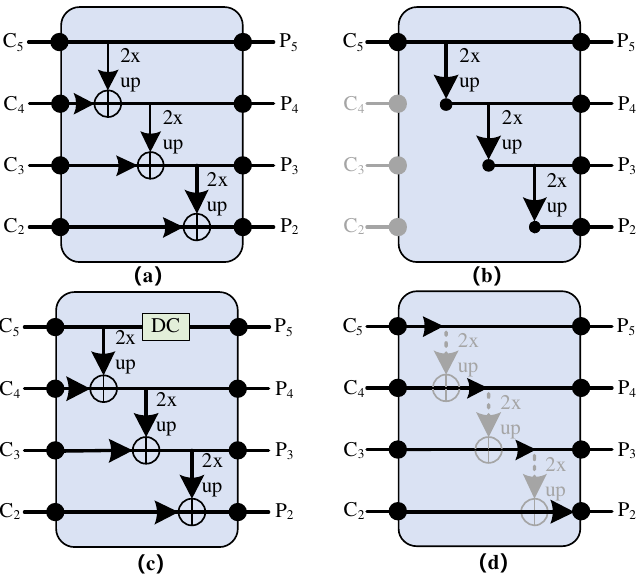}
	\caption{ Illustrations of our experiments of FPN. (a) is the FPN baseline. The single input $C_5$ is acted as an example shown in (b). \textit{DC} denotes the dilated convolution in (c). (d) represents the pyramidal structure without interaction. }
	\label{revisit fpn}
\end{figure}

Before proposing our multi-scale transformer, in this section, we revisit FPN and transformer, exploring the factors that influence the performance of dense multi-scale prediction. Specifically, we first disassemble FPN and examine the effectiveness of each component. Then the transformer is analyzed to discuss the adaptability on multi-scale integration. Based on these analysis, a number of corresponding findings are obtained to give insight into dense multi-scale detection. Here, the experiments are implemented with Faster R-CNN and RetinaNet based on ResNet-50 in object detection task, as shown in Fig \ref{revisit fpn}. Note that $ \left\lbrace C_i \right\rbrace_{i=2}^{5}$ represent the output features from different stages with a down-sampling rate of $\left\lbrace 4, 8, 16, 32 \right\rbrace$ and $ \left\lbrace P_i \right\rbrace_{i=2}^{5}$ denote the features for final prediction. RetinaNet is implemented without the use of $C_2$.

\noindent{\textbf{\textit{Finding 1}: \textit{The high semantic level ($C_5$) plays a significant role on dense multi-scale prediction performance. Single input of $C_5$ can even achieve a comparable accuracy.}}} 

\begin{figure}[!t]
	\centering
	\includegraphics[scale=0.6]{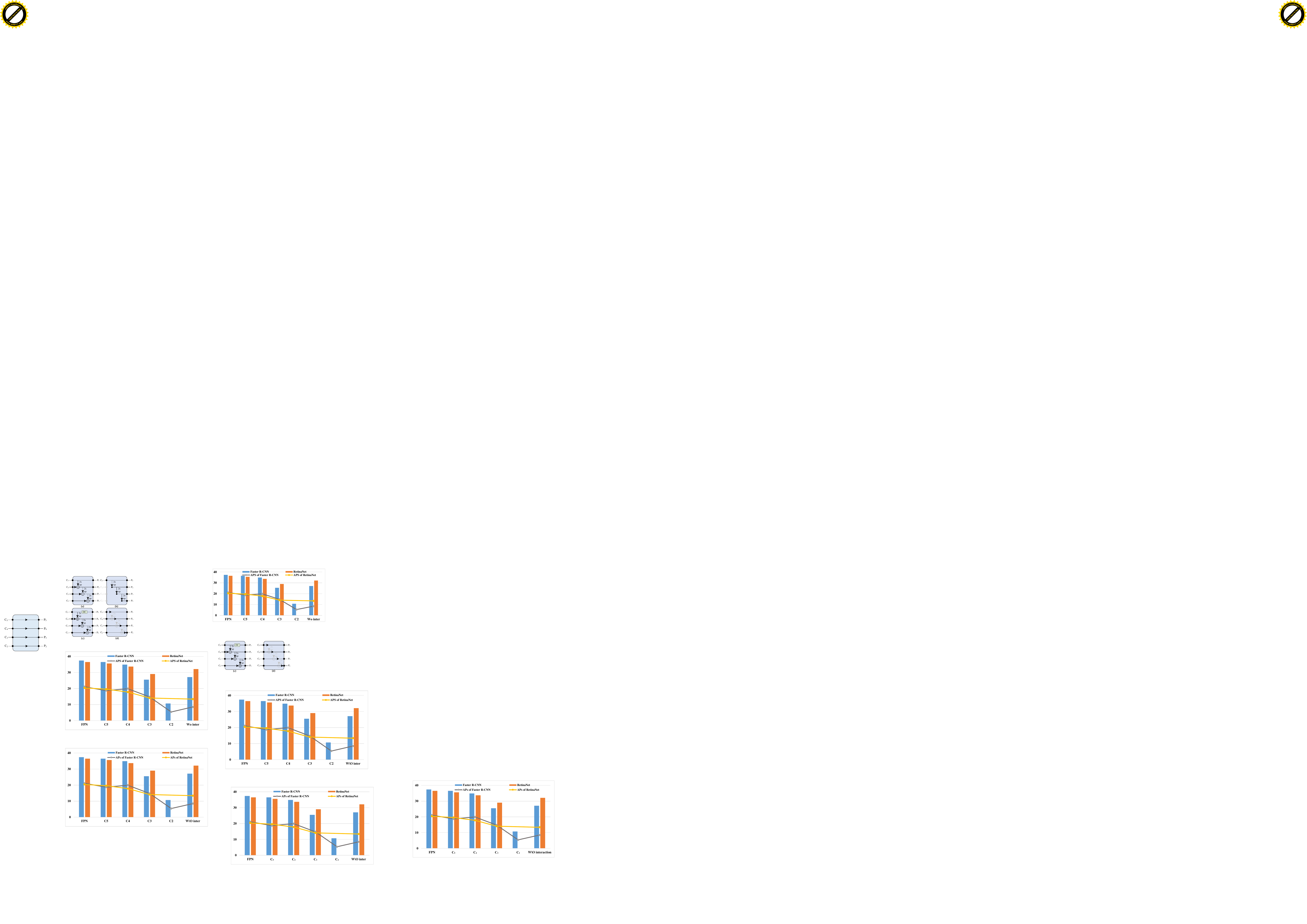}
	\caption{Experimental results in analyzing factors of FPN with Faster R-CNN and RetinaNet on COCO.  $AP_S$ denotes the the average precision of small objects.}
	\label{revisit2}
\end{figure}

As shown in Fig \ref{revisit fpn} (b), we select different levels as singe input of FPN and keep the multi-level output through the operations of down-sampling and up-sampling, in order to validate the importance of different level input features. The experimental results are reported in Fig \ref{revisit2}, which shows that, the single input from all levels has different degrees of performance degradation in Faster R-CNN and RetinaNet likewise has the same phenomenon. Among them, interestingly, model with single $C_5$ input only has slightly accuracy degradation, which is comparable to the baseline with FPN. This observation suggests that $C_5$ has sufficient semantic information, which is vital for detection performance. Based on this, we further enhance $C_5$ by embedding multiple receptive field, which leads to high performance gain. As shown in Fig \ref{revisit fpn} (c), dilated convolutions with a rate of 3 is utilized to enhance the receptive field of $C_5$. Thanks to the extension of receptive field, the model achieves better performance by 0.3 and 0.2 points on Faster R-CNN and RetinaNet. This indicates that adopting multi-receptive field can help semantic level to complementally learn abundant semantic scale information, expanding the diversity of semantic features.

\noindent{\textbf{\textit{Finding 2}: \textit{Although FPN with input of single high semantic level achieves good performance, the interaction among multiple levels is indispensable.}}} 

As presented in Figure \ref{revisit fpn} (d), we directly remove the interaction among different levels to explore the necessity. The results in Fig \ref{revisit2} show that the detection performance has approximate 11 \textit{mAP} reduction in Faster R-CNN and about 5 \textit{mAP} in RetinaNet. Particularly, the small objects' performance of both detectors has steep decrease. Furthermore, although single input of $C_5$ can achieve the comparable result with baseline, the performance of four levels' input drops dramatically in both detectors without interaction. Thus, these phenomenons indicate that: (1) the interaction among different levels is indispensable to fully exploit the complementary roles of each feature level for dense multi-scale prediction; (2) there exists redundant information in low levels affecting the performance.

\begin{table}[!t]
	\centering
	\caption{Results of different attempts of applying transformer to FPN. }
	\resizebox{0.4 \textwidth}{!}{
		\begin{tabular}{|c|c|c|c|c|c|c|}
			\hline
			Method& $AP$ & $AP_S$& $AP_M$&$AP_L$&Flops (G)\\
			\hline
			baseline &37.4&21.2&41.0&48.1&207.07\\
			\hline
			p-MSA  &-&-&-&-&-\\
			
			s-MSA &37.3&19.9&41.7&49.7& 229.87   \\
			d-MSA &37.7&22.1&41.5&48.5&218.46   \\
			
			\hline
	\end{tabular}}
	\label{attempt}
\end{table}

\begin{figure*}[!t]
	\centering
	\includegraphics[scale=0.45]{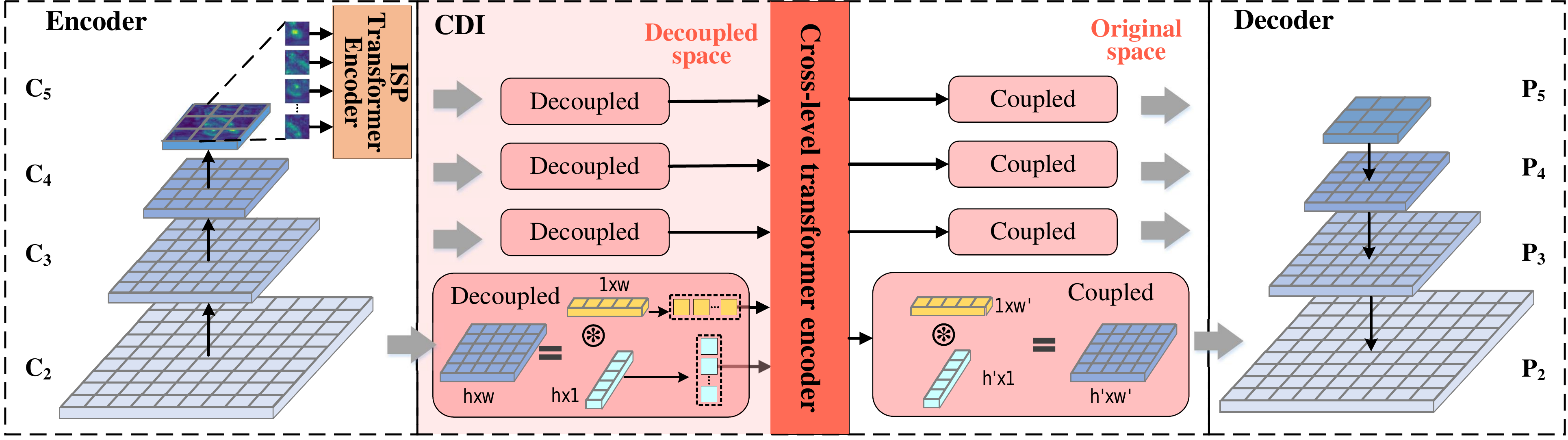}
	\caption{The framework of the proposed SDTP.}
	\label{pipe}
\end{figure*}

\noindent{\textbf{\textit{Finding 3}: \textit{Feature interaction in the proposed decoupled space is promising to promote the efficiency of transformer with high performance.}}}

Transformer is good at capturing the global relationship by multi-head self attention (MSA), but there is no suitable method to solve the scale variation problem in dense image prediction tasks. Here, we conduct three attempts to apply transformer to multi-scale detection: (1) primitive MSA (p-MSA): the primitive transformer block is directly applied to different levels of FPN; (2) stride-MSA (s-MSA): in order to reduce the computation, ($C_2$, $C_3$, $C_4$) are down-sampled with strides of (8, 4, 2) to achieve the same size with $C_5$, respectively; (3) decoupled MSA (d-MSA): we propose a decoupled style of features \footnote{We use pooling operator to decouple $h$ and $w$.} to represent features with low flops. Given an image containing $h \times w$ patches with $c$ channels, the computational complexity of three attempts can be described as:

\begin{equation}
\small
\mathcal{O}(p\verb|-|MSA)=\sum_{i=2}^{5}(4h_iw_ic_{i}^2+2(h_iw_i)^2c_i) ,
\end{equation}
\begin{equation}
\small
\mathcal{O}(s\verb|-|MSA)=\sum_{i=2}^{5}(4\frac{h_iw_i}{s_{i}^2}c_{i}^2+2(\frac{h_iw_i}{s_{i}^2})^2c_i) ,
\end{equation}
\begin{equation}
\small
\mathcal{O}(d\verb|-|MSA)=\sum_{i=2}^{5}(4(h_i+w_i)c_{i}^2+2(h_i^2+w_i^2)c_i).
\end{equation}
As shown in Tab \ref{attempt}, it can be seen that the implementation of p-MSA fails due to the huge computation cost. s-MSA and d-MSA can significantly decrease the heavy computation, which verifies the efficiency of s-MSA and d-MSA. However, s-MSA performs worse than d-MSA and d-MSA, which has the smallest computation, even obtains the best performance. This may be because features of low levels are generally of large spatial size and contain redundant information, while our decoupled attempt can achieve computational savings with better performance and remove some redundancy. Thus, the decoupled-style method gives us some enlightenments to design an efficient and effective transformer to enable interaction among representative features. 

\section{The Proposed Method}
According to our above findings, diverse and sufficient semantic information, effective multi-scale interaction and proper feature dimension reduction are all important for applying transformer in a multi-scale way. Hence, we propose a novel and effective method called SDTP as depicted in Fig \ref{pipe}. In the proposed SDTP framework, encoder extracts multi-scale features and decoder transmits the semantic and scale information from high level to low level. Besides, three main components, including ISP, CDI and ARF, are designed to apply multi-scale technique to transformer. In particular, ISP transformer encoder is applied to the high semantic level ($C_5$) for exploring intra-level semantic diversity. CDI adopts the decoupled style of features to effectively enable the sufficient interaction among tokens from various levels by cross-level transformer encoder. Besides, ARF is embedded in the attention module of transformer block for precise correlation result.    

\subsection{Intra-level Semantic Promotion (ISP)} 
Transformer is proficient in capturing global information relying on its core module self-attention. However, self-attention generally acts on the single feature to build its long range information. This mechanism only focuses on the current features' states, which ignores to explore the feature diversity (e.g., multi-receptive field of one feature). Even though in pyramid structure, high semantic feature has strong representation, but still losses diverse semantic scale detail information. As proved in finding 1, we learn that the semantic level (i.e., $C_5$) contains useful context which is crucial to performance, and the enhancement of $C_5$ (i.e., multi-receptive enhancement) can obtain a satisfied performance gain. Thus, we design the ISP transformer encoder to explore semantic diversity information of the high semantic level in receptive space, which can flexibly integrate the local and global information in the transformer.

\begin{figure}[ht]
	\centering
	\includegraphics[scale=0.43]{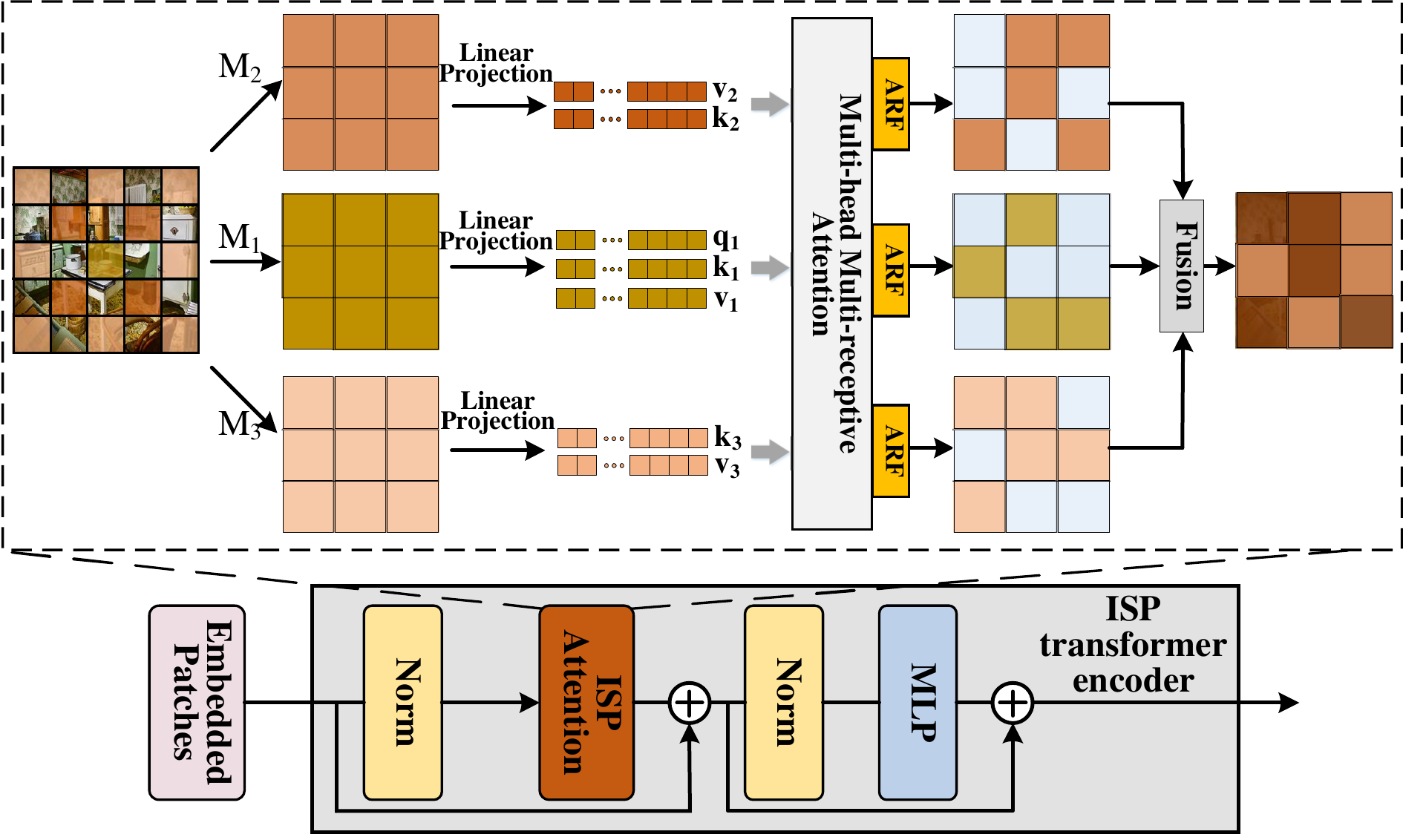}
	\caption{Illustration of ISP transformer encoder. $ARF$ represents the attention refinement function module.}
	\label{isp}
\end{figure}

In particular, as shown in Fig \ref{isp}, the whole process of ISP can be formulated as:
\begin{equation}
\small
\hat{C_5}=Attn_{\mathrm{ISP}}(LN(C_5))+C_5,
\end{equation}
\begin{equation}
\small
{C_5^\star}=MLP(LN(\hat{C_5}))+\hat{C_5},
\end{equation}
where $LN (\cdot)$ denotes the normalization layer and $MLP (\cdot)$ is multi-layer perception. $Attn_{\mathrm{ISP}} (\cdot)$ is the key module to enhance the semantic diversity through multi-receptive tokens, which can be written as:  
\begin{equation}
\small
Attn_{\mathrm{ISP}}(C_5) =MMA[\mathcal{G}(C_5)],
\end{equation}
in which $\mathcal{G}(\cdot)$  Eq. \ref{gmrt} is the function to generate multi-receptive tokens and $MMA (\cdot)$ denotes the multi-head multi-receptive attention in Eq. \ref{mma}. We explain these two components in the following:
\begin{equation}
\small
\mathcal{G}(C_5)={\left\lbrace q_{s}, k_s, v_s\right\rbrace}_{s=1}^{S} ={\left\lbrace Reshape(M_s)\right\rbrace}_{s=1}^{S} \in \mathbb{R}^{c\times (hw)},
\label{gmrt}
\end{equation}
\begin{equation}
\small
s.t. \quad  M_{s} =\sum\limits_
{\Phi\in \mathscr{O}_s}\Phi(C_5),\quad  s=(1,...S).
\end{equation}
$M_s\in \mathbb{R}^{c\times h\times w}$ denotes the various states of features with different receptive fields. ${\mathscr{O}}_s$ represents the operation set, including $3 \times 3$ dilated convolution with different rates and the position embedding. $S$ represents the number of dilated rates which we adopt $3$ in our method. To keep the original local information without loss, we always include rate of 1. Obtaining query $q$, key $k$ and value $v$ as shown, we utilize the initial state' query $q_1$ to explore and search complementary semantic scale information of other states by $MMA(\cdot)$. Details can be formulated as follows:
\begin{equation}
\small
MMA=\mathbb{C} [\left\lbrace Attn(q_{1}w_{d}^{q},k_{s}w_{d}^{k},v_{s}w_{d}^{v})\right\rbrace_{d=1}^{D}],\quad  s=(1,2,3).
\label{mma}
\end{equation}
Here, $\mathbb{C}(\cdot)$ means concatenation and $D$ is the number of heads. $Attn (\cdot)$ computes the token-wise correlation among the inputs. $w_{j}^{q},w_{j}^{k},w_{j}^{v}$ are the linear projection parameters. Our design keeps the query $q_1$ of high priority to avoid the information loss. And we utilize $q_1$ to search the diverse semantic message in a local-to-global way, which helps mine the diversity and make up the deficiency. Besides, due to the tiny spatial size of $C_5$, the computational cost is limited.

In summary, we realize intra-level diversity semantic promotion due to the ISP transformer encoder, which aggregates multi-receptive information in a local-to-global way. ISP makes full use of the superiority and diversity of high semantic level, which can flexibly explore its different scale states to search effective information in receptive space.

\subsection{Cross-level Decoupled Interaction (CDI)}
Transformer has a dominant performance in NLP task due to capturing long-range relationship. Hence, it is desirable to establish cross-level interaction based on transformer. However, there exist two fatal obstacles. The first one is huge computation cost. In transformer-based vision tasks, we regard image patches as tokens, and the resulted sequence length is generally large. What is worse, comparing with classification task, dense image prediction tasks like object detection need image of higher-resolution to gain precise prediction. Thus, applying transformer on features of high resolution in dense prediction tasks is difficult, let alone the interaction among different levels. Secondly, the popular interaction style in FPN is insufficient and rigid, since non-adjacent levels can not learn from each other. To address these problems, we propose the cross-level decoupled interaction module (CDI).

On the one hand, it reduces the input dimension for transformer in decoupled space, making multi-scale interaction practical and efficient. On the other hand, the proposed decoupled operation effectively decrease the redundancy in original features, which thereafter deepen the multi-scale interaction and improves the performance.

\begin{figure}[t]
	\centering
	\includegraphics[scale=0.45]{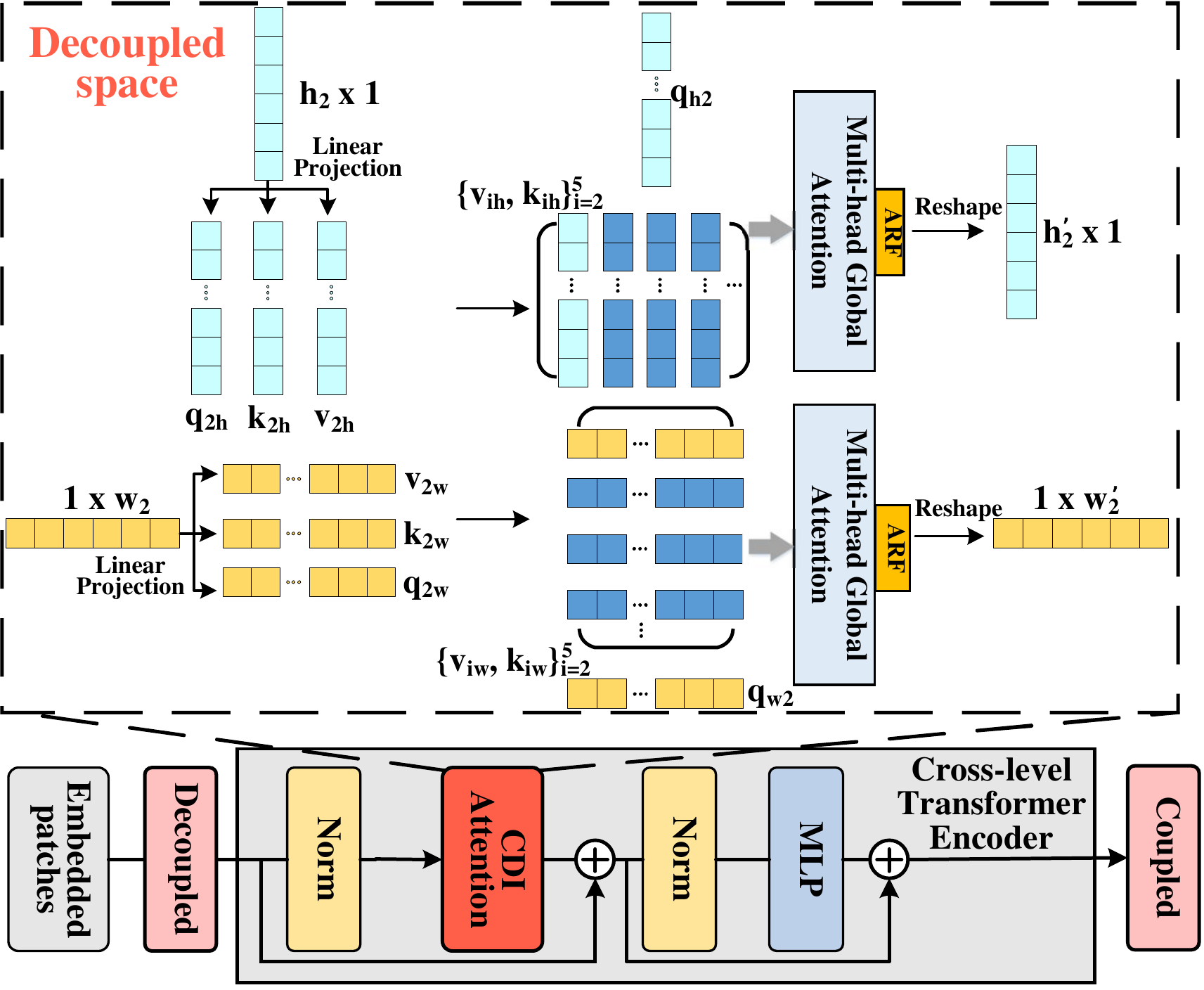}
	\caption{Illustration of cross-level transformer encoder. We use $C_2$ as an example to show the process.}
	\label{cdi}
\end{figure}

As shown in Fig \ref{pipe}, the proposed CDI decouples one feature map into a the vertical feature and a horizontal one with much smaller dimensions. For clearness, we present the decoupled process in Eq. \ref{10} and Eq. \ref{11}.
\begin{equation}
\small
Y_i=\mathcal{F}_{3*1}[\sum_{j}^{w} \Psi [\mathcal{F}_{1*1}(C_i)] \cdot \sum_{j}^{w}(C_i)] \in \mathbb{R}^{c\times h \times 1},
\label{10}
\end{equation}
\begin{equation}
\small
X_i=\mathcal{F}_{1*3}[\sum_{j}^{h} \Psi [\mathcal{F}_{1*1}(C_i)] \cdot \sum_{j}^{h}(C_i)] \in \mathbb{R}^{c\times 1 \times w},
\label{11}
\end{equation}

in which $\Psi$ denotes the activation function and $\mathcal{F}$ is the convolution operations with its kernel size shown in the subscript to enhance the learnability. $Y_i$ and $X_i$ are the decoupled features of $C_i$. 
Considering the decoupled features from multiple levels as tokens, we leverage them to make a flexible and sufficient interaction in cross-level transformer as shown in Fig \ref{cdi}. Firstly, we can obtain $q, k, v$ from $Y_{i}$ and $X_{i}$:
\begin{equation}
\small
\left\lbrace q_{ih}, k_{ih}, v_{ih}\right\rbrace_{i=2}^{5}=\left\lbrace Reshape( Y_{i})\right\rbrace_{i=2}^{5} \in \mathbb{R}^{c\times h},
\end{equation}
\begin{equation}
\small
\left\lbrace q_{iw}, k_{iw}, v_{iw}\right\rbrace_{i=2}^{5}=\left\lbrace Reshape( X_{i})\right\rbrace_{i=2}^{5}\in \mathbb{R}^{c\times w}.
\end{equation}
Then we design a multi-head global attention module (MGA) to make the features learn cross-level knowledge from each other. Note that the MGA for tokens from decoupled vertical and horizontal features are implemented separately. For tokens form $X_i$, this process can be formulated as below, and the MGA for $Y_i$ is implemented similarly.
\begin{equation}
\small
MGA=\mathbb{C} [ Attn(q_{i}w^{q},\left\lbrace k_{i}w^{k}\right\rbrace_{i=2}^{5} ,\left\lbrace v_{i} w^{v}\right\rbrace_{i=2}^{5})]_{d=1}^{D}.
\end{equation}
The output of MGA are $\hat{X_i}$ and $\hat{Y_i}$ with sufficient scale information from all levels owing to the sufficient interaction. Particularly, in contrast to the traditional top-down fusion style in FPN, our design is more flexible since it allows for interactions for features of any two levels. Finally, we make the re-couple of $\hat{C_i}$ for the last process such as $MLP$ in transformer block.

Besides, in order to avoid significant information loss, we introduce an effective loss function for decoupled process to achieve the end-to-end optimization:
\begin{equation}
\small
\mathcal{L}_{dep}=\sum_{i=2}^{5}||C_i-(Y_i \circledast X_{i})||_2 .
\end{equation}
Note that $\circledast$ is the Kronecker sum operator. The total loss can be described as:
\begin{equation}
\small
\mathcal{L}=\mathcal{L}_{org}+\lambda	\mathcal{L}_{dep} ,
\end{equation}
in which $\mathcal{L}_{org}$ represent the original loss of dense prediction task. Besides, we leverage $\lambda$ to balance the loss, which is set to 0.01 in our experiments.

Because of the large size and redundant information in lower levels of dense prediction tasks, applying transformer to capture long-range relationship is impractical. The proposed CDI adopts the decoupled style to obtain more representative features with low dimension, and makes tokens learn adequate cross-level message through the multi-head global attention, which proves to be efficient and effective.

\subsection{Attention Refinement Function (ARF)}
As the core block in transformer, self-attention module generates an attention map that depicts the correlation between different tokens. The previous studies (e,g., \cite{dosovitskiy2020image, wang2021pyramid}) adopt Softmax as the activation function in correlation calculation. However, Softmaxt takes account of all tokens together, ignoring the independence and inconsistency of various tokens. In dense prediction tasks, there are enormous tokens and the use of Softmax is likely to alter the correlation map by enhancing some relations and weakening others. However, we desire for a correlation map that can reflect the relations more accurately.

To alleviate this problem, we propose a flexible and effective activation function, which is formulated as:
\begin{equation}
\small
\mathcal{U}(x)=Max[\frac{e^x-e^{-x}}{e^x+e^{-(x+2\tau)}},0],
\label{activation}
\end{equation}
where, $\tau$ is a hyper-parameter. Our proposed activation function considers each correlation individually and eliminates the redundant relations, avoiding the limitations of Softmax. Moreover, it is easy to find that Tanh is a special case of this function when the hyper-parameter $\tau$ is set to zero and the activation value is positive. Comparing with Tanh, this function can boost values that are close to zero by adjusting the hyper-parameter $\tau$ and suppress the redundant relations to zero whose activation values are negative. Rather than using Softmax as activation function, we rethink the calculation of attention in self-attention and propose ARF to refine the correlation result. ARF can be put in other transformer-based methods as a pluggable component to refine the attention.

\section{Experiments}
\subsection{Settings}
\noindent{\textbf{Data and evaluation.}}
Our experiments are implemented on the MS COCO 2017 \cite{lin2014microsoft} for object detection and instance segmentation and ADE20K \cite{zhou2017scene} for semantic segmentation. MS COCO 2017 contains 80 object categories for detection and instance segmentation tasks, which consists of 115k images for training \textit{(train2017)} and 5k images for validation \textit{(val2017)}. We train on \textit{train2017}, and report results on \textit{val2017}. Then performance is evaluated by standard COCO-style Average Precision (AP) metrics on small, medium and large objects, i.e., AP$_s$, AP$_m$ and AP$_l$. Besides, AP$^b$ and AP$^m$ denote the AP of bounding box and mask. ADE20K is a challenging scene parsing benchmark containing 150 fine-grained semantic categories , consisting of 20k images for training and 2k images for validation. The mean Intersection-over-Union (mIoU) is the primary metric to evaluate the performance of semantic segmentation.

\noindent{\textbf{Implementation details.}}
We implement our method based on mmdetection \cite{chen2019mmdetection}. In order to ensure the fairness of comparisons, we also re-implement baseline methods on mmdetection \cite{chen2019mmdetection}. The results in our re-implementation are generally better than those in the reference papers. Besides, the backbone used in our experiments are pre-trained in ImageNet \cite{deng2009imagenet}. If not otherwise specified, all the baseline methods are equiped with FPN and the ablation studies are implemented with Faster R-CNN based on ResNet50. All other hyper-parameters in our work follow the settings in mmdetection. 
\subsection{Performance}
To demonstrate the generality of our method, we implement our framework on three dense prediction tasks: object detection, semantic segmentation and instance segmentation.

\noindent{\textbf{Object detection.}}
The results on common object detectors are shown in Tab \ref{detection}. The proposed SDTP achieves consistent improvement on both single-stage and two-stage detectors. When paired with strong detectors, SDTP still shows the superiority with better performance. In particular, the results of small objects has significant improvement due to the effective utilization of high-resolution features because of the sufficient interaction among multiply levels.
\begin{table}
	\centering
	\caption{\textbf{Object Detection:} Performance comparisons with popular detectors. "SDTP" denotes our method. ``$\surd$" means the baselines integrated with our transformer pyramid.}
	\resizebox{0.45\textwidth }{!}{
		\begin{tabular}{|c|c|c|cccc|}
			\hline
			Method&Backbone&SDTP&  $AP^b$ & $ AP^b_{S}$ & $AP^b_{M}$ & $AP^b_{L}$ \\
			\hline
			\multirow{4}{*}{RetinaNet}&\multirow{2}{*}{R50} & &36.5&20.4&40.3&48.1\\
			& &$\surd$&\textbf{38.1}&\textbf{21.8}&\textbf{41.8}&\textbf{49.1}  \\
			\cline{2-7}                      
			&\multirow{2}{*}{R101}&&38.5&21.7&42.8&50.4    \\
			&&$\surd$&\textbf{40.0}&\textbf{22.5}&\textbf{44.1}&\textbf{52.1} \\
			
			\hline
			\hline
			\multirow{4}{*}{Faster R-CNN}&\multirow{2}{*}{R50} & &37.4&21.2&41.0&48.1\\
			
			& &$\surd$&\textbf{39.4}&\textbf{22.7}&\textbf{42.7}&\textbf{51.0}\\
			\cline{2-7}                      
			&\multirow{2}{*}{R101}&&39.4&22.4&43.7&51.1\\
			
			&&$\surd$&\textbf{40.8} & \textbf{23.3} & \textbf{44.9} & \textbf{54.0} \\
			
			\cline{2-7}
			
			\hline
			\hline
			
			\multirow{4}{*}{Cascade R-CNN}&\multirow{2}{*}{R50} & &40.3&22.5&43.8&52.9\\
			& &$\surd$&\textbf{41.7}&  \textbf{24.2}& \textbf{45.0} & \textbf{54.9}\\
			\cline{2-7}
			&\multirow{2}{*}{R101}&&42.0&23.4&45.8&55.7\\
			&&$\surd$&\textbf{43.2}&\textbf{25.3}&\textbf{47.1}&\textbf{57.3} \\
			\hline
			
	\end{tabular}}
	\label{detection}
\end{table}

\noindent{\textbf{Semantic segmentation.}}
We also conduct experiments to prove the effectiveness of SDTP on semantic segmentation task. As shown in Tab \ref{semantic seg}, we compare our method with Semantic FPN \cite{kirillov2019panoptic} and PointRend. With the help of SDTP, we again outperform the baselines. Especially, the $mIoU$ of PointRend with ResNet50 increases nearly 4 points with SDTP. Since the semantic segmentation is more sensitive to multi-resolution information, the performance margin brought by SDTP is remarkable.

\begin{table}[!h]
	\centering
	\caption{\textbf{Semantic Segmentation:} Performance comparisons with common semantic segmentation methods of different backbones on ADE20K validation set.}
	\resizebox{0.45\textwidth }{!}{
		\begin{tabular}{|c|c|c|ccc|}
			\hline
			Method&Backbone&SDTP&  $mIoU$ & $ mAcc$ & $aAcc$  \\
			\hline
			\multirow{4}{*}{Semantic FPN}&\multirow{2}{*}{R50} & &37.48&47.57&78.02\\
			& &$\surd$&\textbf{38.77}&\textbf{49.35}&\textbf{79.13}  \\
			\cline{2-6}                      
			&\multirow{2}{*}{R101}&&39.35&49.43&79.19   \\
			&&$\surd$&\textbf{41.52}&\textbf{51.68}&\textbf{80.13} \\
			
			\hline
			\hline
			\multirow{4}{*}{PointRend}&\multirow{2}{*}{R50} & &37.63&48.14&77.80\\
			
			& &$\surd$&\textbf{41.53}&\textbf{51.82}&\textbf{79.56}\\
			\cline{2-6}                      
			&\multirow{2}{*}{R101}&&40.01&50.56&79.09  \\
			&&$\surd$&\textbf{42.39}&\textbf{53.44}&\textbf{80.35} \\	
			
			\hline

	\end{tabular}}
	\label{semantic seg}
\end{table}

\begin{table*}
	\centering
	\caption{\textbf{Comparisons with the state-of-the-art methods:} The symbol “*” means our re-implemented results on mmdetection.}
	\resizebox{0.9 \textwidth}{!}{
		\begin{tabular}{|c|cc|cccccc|}
			
			\hline
			Method  & Backbone & Schedule & $AP$ & $AP_{50}$ & $AP_{75} $&$ AP_{S}$ & $AP_{M}$ & $AP_{L}$ \\
			\hline
			Faster R-CNN*        & ResNeXt101-32$\times$4d&12  & 41.2 & 62.1 & 45.1 & 24.0 & 45.5 & 53.5       \\
			Faster R-CNN*    & ResNeXt101-64$\times$4d  &12 & 42.1 & 63.0 & 46.3 & 24.8 & 46.2 & 55.3      \\
			Mask R-CNN*    & ResNeXt101-32$\times$4d  &12 & 41.9 & 62.5 & 45.9 & 24.4 & 46.3 & 54.0      \\
			Cascade R-CNN*    & ResNeXt101-32$\times$4d&12   & 43.7 & 62.3 & 47.7 & 25.1 & 47.6 & 57.3 \\
			DETR\cite{carion2020end}   & ResNet50&500  & 42.0 & 62.4 & 44.2 & 20.5 & 45.8 & 61.1 \\
			DETR\cite{carion2020end}   & ResNet101&500 & 43.5 & 63.8 & 46.4 & 21.9 & 48.0 & \textbf{61.8} \\
			Deformable DETR\cite{zhu2020deformable}   & ResNet50&50 & 43.8 & 62.6 & 47.7 & 26.4 & 47.1 & 58.0 \\
			Sparse R-CNN\cite{sun2020sparse}   & ResNet101 &36  & 44.1 & 62.1 & 47.2 & 26.1 & 46.3 & 59.7      \\
			Cascade Mask R-CNN*   & ResNeXt101-32$\times$4d &20  & 45.0 & 63.2 & 49.1 & 26.7 & 48.9 & 59.0      \\
			HTC*                              & ResNet101 &20  & 44.8 & 63.3 & 48.8 & 25.7 & 48.5 & 60.2 \\
			
			\hline
			\hline
			SDTP Faster R-CNN(ours)   & ResNeXt101-32$\times$4d  & 12 & 42.3 & 63.7 & 46.1 & 25.3 & 46.6 & 54.8   \\
			SDTP Faster R-CNN(ours)      	  & ResNeXt101-64$\times$4d&12   & 43.0 & 63.9 & 46.6 & 25.3 & 46.9 & 55.8\\
			SDTP Mask R-CNN(ours)  & ResNeXt101-32$\times$4d &12  & 43.2 & 64.3 & 47.1 & 25.9 & 47.1 & 56.5\\
			SDTP Cascade R-CNN(ours)   & ResNeXt101-32$\times$4d &12 & 44.6 & {63.8} & {48.6} & {26.1} & {48.7} & 57.8\\
			SDTP Sparse R-CNN(ours)   & ResNet101 &36 & 44.6 &{63.1} & {48.3} & {27.3} & {47.6} & 60.0\\
			SDTP Cascade Mask R-CNN (ours)   & ResNeXt101-32$\times$4d &20  & 45.7 & 64.5 & 49.8 & 27.1 & 49.1 & 59.6      \\
			SDTP HTC* (ours)                             & ResNet101 &20  & \textbf{45.8} & \textbf{64.9} & \textbf{49.8} & \textbf{27.3} & \textbf{49.3} & 60.5 \\
			\hline
	\end{tabular}}
	\label{sota}
\end{table*}
\noindent{\textbf{Instance segmentation.}}
We continue to validate the generalization of SDTP on instance segmentation task as shown in Tab \ref{instance}. Our method improves the baseline on both detection and instance segmentation task with great margin. Even with the strong method such as HTC, SDTP still has significant increase by 1.5 points. Besides, benefiting from the diverse semantic information, the performance on large object in instance segmentation has witnessed dominant promotion.
\begin{table}[!ht]
	\centering
	\caption{\textbf{Instance Segmentation:} Performance comparisons with strong instance segmentation methods.}
	\resizebox{0.45\textwidth }{!}{
		\begin{tabular}{|c|c|c|cccc|}
			\hline
			Method&Backbone&SDTP&  $AP^{b}$ & $ AP^{b}_{S}$ & $AP^{m}$ & $AP^{m}_{L}$ \\
			\hline
			\multirow{4}{*}{Mask R-CNN}&\multirow{2}{*}{R50} & &38.2&21.9&34.7&47.2\\
			& &$\surd$&\textbf{40.0}&\textbf{22.8}&\textbf{36.2}&\textbf{53.1}  \\
			\cline{2-7}                      
			&\multirow{2}{*}{R101}&&40.0&22.6&36.1&49.5   \\
			&&$\surd$&\textbf{41.6}&\textbf{24.5}&\textbf{37.2}&\textbf{54.7} \\
			
			\hline
			\hline
			\multirow{2}{*}{PointRend}&\multirow{2}{*}{R50} & &38.4&22.8&36.3&48.5\\
			
			& &$\surd$&\textbf{40.9}&\textbf{25.5}&\textbf{38.0}&\textbf{50.8}\\

			\hline
			\hline            
			\multirow{2}{*}{HTC}&\multirow{2}{*}{R50} & &42.3&23.7&37.4&51.7\\
			
			& &$\surd$&\textbf{43.8}&\textbf{25.7}&\textbf{38.9}&\textbf{57.2}\\
			

			\hline

	\end{tabular}}
	\label{instance}
\end{table}

\noindent{\textbf{Comparison on transformer-based method.}}
Apart from the above experiments, we further assess the superiority and generality of SDTP on transformer-based backbone. As seen in Tab \ref{transformer}, we apply our method to Mask R-CNN based on backbones of two versions of PVT. SDTP still obtains better performance on two dense prediction tasks.
\begin{table}[!h]
	\centering
	\caption{\textbf{Comparison with transformer-based backbone:} Performance comparisons paired with Mask R-CNN.}
	\resizebox{0.45\textwidth }{!}{
		\begin{tabular}{|c|c|c|cccc|}
			\hline
			Method&Backbone&SDTP&  $AP^b$ & $ AP^b_{S}$ & $AP^{m}$ & $AP^{m}_{L}$ \\
			\hline
			\multirow{4}{*}{Mask R-CNN}&\multirow{2}{*}{PVT-Tiny} & &36.7&21.6&35.1&48.5\\
			& &$\surd$&\textbf{38.3}&\textbf{23.5}&\textbf{36.2}&\textbf{54.2}  \\
			\cline{2-7}                      
			&\multirow{2}{*}{PVT-Small}&&40.4&22.9&37.8&53.6   \\
			&&$\surd$&\textbf{41.4}&\textbf{23.3}&\textbf{38.3}&\textbf{58.0} \\ 
			
			\hline
			
	\end{tabular}}
	\label{transformer}
\end{table}

\noindent{\textbf{Comparison with State-of-the-art methods.}}
All shown in Tab \ref{sota}, SDTP achieves consistently non-negligible improvements even with more powerful backbones and more training epochs. For example, when applying ResNeXt101-32$\times$4d and ResNeXt101-64$\times$4d as the feature extractors of Faster R-CNN, our SDTP still improves performance by 1.1 and 0.9 points, respectively. Besides, in the comparison between HTC and its SDTP variant which are both trained with 20 epochs, SDTP still wins the comparisons. SDTP brings consistent improvements on various backbones, methods and learning schedules. This proves the generalization and superiority of SDTP.

\subsection{Ablation studies}
\noindent{\textbf{Ablation studies on each component.}} To analyze the importance of each module in STDP, we gradually apply them to the model. As in Tab \ref{each}, all three parts are essential. Especially, we can find that ISP alone increases 1.3 points. The adding of CDI helps to give a boost of 1.0 point. These results are in consistency with our first two findings that both semantic diversity and interaction across multi-level features are the core of multi-scale technique. Furthermore, the proposed ARF provides more accurate attention map in building the intra-level and cross-level transformer, and exploiting ARF enables the performance grow to 39.4. 
\begin{table}[!h]
	\centering
	\caption{\textbf{Effectiveness of each component.}}
	\resizebox{0.45 \textwidth}{!}{
		\begin{tabular}{|c|cccc|}
			\hline
			Method& $AP$ & $ AP_{S}$ & $AP_{M}$ & $AP_{L}$ \\
			\hline
			baseline  &37.4&21.2&41.0&48.1\\
			\hline
			baseline+ISP &38.7&22.3&42.5&50.8    \\
			baseline++CDI  & 38.4&21.3&42.0&49.9      \\
			baseline+ISP+CDI  & 39.0&22.6&42.6&50.7      \\
			baseline+ISP+CDI+ARF  &  \textbf{39.4}&\textbf{22.7}&\textbf{42.7}&\textbf{51.0}    \\
			\hline
	\end{tabular}}
	\label{each}
\end{table}

\noindent{\textbf{Ablation studies on different dilated rates in ISP.}} 
The results illustrated in Tab \ref{ISP} show that exploring receptive space brings improvements to SDTP. In detail, the combination of 1,3,6 has the best performance. Another observation is that the performances of the first two rows are lower, which indicates that the selected dilation rates should be of large difference so that the semantic features can be diverse.
\begin{table}[!ht]
	\centering
	\caption{\textbf{Different settings of dilated rates in ISP.}}
	\resizebox{0.30 \textwidth}{!}{
		\begin{tabular}{|c|cccc|}
			\hline
			Rates& $AP$ & $ AP_{S}$ & $AP_{M}$ & $AP_{L}$ \\
			\hline
			1,2,3  &38.1&21.5&41.6&49.6\\
			
			1,2,4 &37.9&21.4&41.2&49.1    \\
			\textbf{1,3,6}  & \textbf{38.7}&\textbf{22.3}&\textbf{42.5}&\textbf{50.8 }     \\
			2,4,6  & 38.4&21.9&42.2&49.1      \\
			
			3,6,12  &  {38.3}&{21.4}&{41.6}&{50.0}    \\
			\hline
	\end{tabular}}
	\label{ISP}
\end{table}

\noindent{\textbf{Ablation studies on $\tau$ in ARF.}} Experimental results related with the settings of ARF are presented in Tab \ref{ARF}. Paired with our proposed function, the suppression of irrelevant information helps for better performance. We choose 2 for $\tau$ which achieves best performance. 
\begin{table}[!ht]
	\centering
	\caption{\textbf{Comparisons among different $\tau$ of ARF}.}
	\resizebox{0.33 \textwidth}{!}{
		\begin{tabular}{|c|c|cccc|}
			\hline
			Method&$\tau$& $AP$ & $ AP_{S}$ & $AP_{M}$ & $AP_{L}$ \\
			\hline
			Softmax &- &39.0&22.6&42.6&50.7\\
			
			Tanh &-&39.1&22.4&42.6&51.0    \\
			\hline
			\multirow{4}{*}{Proposed} & 1& 39.0&22.1&42.5&50.9      \\
			&\textbf{2} & \textbf{39.4}&\textbf{22.7}&\textbf{42.7}&{51.0 }     \\
	
			&3 &  {39.1}&{22.4}&{42.2}&{50.8}    \\
			&4 &  {39.2}&{22.6}&{42.5}&\textbf{{51.4}}    \\
			\hline
	\end{tabular}}
	\label{ARF}
\end{table}
\section{Conclusion}
In this paper, we focus on dealing with the scale variation problem in dense image prediction tasks with the aid of multi-scale and transformer techniques. To begin with, we revisit the dense multi-scale prediction, and obtain important insights that semantic diversity and interaction among different levels are the key elements. Based on these findings, we propose a novel semantic-aware decoupled transformer pyramid, which includes three simple yet effective components, i.e., Intra-level Semantic Promotion, Cross-level Decoupled Interaction and Attention Refinement Function. SDTP has shown the generality and effectiveness on various dense image prediction tasks. Additionally, SDTP and its three key components can be easily extended to other methods.

{
\bibliographystyle{ieee_fullname}
\bibliography{egbib}
}

\end{document}